\documentclass[lettersize,journal]{IEEEtran}
\usepackage{amsmath,amsfonts}
\usepackage{array}
\usepackage[caption=false,font=normalsize,labelfont=sf,textfont=sf]{subfig}
\usepackage{textcomp}
\usepackage{stfloats}
\usepackage{url}
\usepackage{verbatim}
\usepackage{graphicx}
\usepackage{algorithm}
\usepackage{algpseudocode}
\usepackage{amsmath,amssymb}
\usepackage{multirow} 
\usepackage[table]{xcolor}
\usepackage{float}
\def\BibTeX{{\rm B\kern-.05em{\sc i\kern-.025em b}\kern-.08em
    T\kern-.1667em\lower.7ex\hbox{E}\kern-.125emX}}
\usepackage{balance}
\begin{document}
\title{Riemannian and Symplectic Geometry for Hierarchical Text-Driven Place Recognition}
\author{Tianyi Shang and Zhenyu Li*}
%\thanks{Manuscript created October, 2020; This work was developed by the IEEE Publication Technology Department. This work is distributed under the \LaTeX \ Project Public License (LPPL) ( http://www.latex-project.org/ ) version 1.3. A copy of the LPPL, version 1.3, is included in the base \LaTeX \ documentation of all distributions of \LaTeX \ released 2003/12/01 or later. The opinions expressed here are entirely that of the author. No warranty is expressed or implied. User assumes all risk.}}

%\markboth{Journal of \LaTeX\ Class Files,~Vol.~18, No.~9, September~2020}%
%{How to Use the IEEEtran \LaTeX \ Templates}

\maketitle

\begin{abstract}
Text-to-point-cloud localization enables robots to understand spatial positions through natural language descriptions, which is crucial for human-robot collaboration in applications such as autonomous driving and last-mile delivery. However, existing methods employ pooled global descriptors for similarity retrieval, which suffer from severe information loss and fail to capture discriminative scene structures. To address these issues, we propose SympLoc, a novel coarse-to-fine localization framework with multi-level alignment in the coarse stage. Different from previous methods that rely solely on global descriptors, our coarse stage consists of three complementary alignment levels: 1) Instance-level alignment establishes direct correspondence between individual object instances in point clouds and textual hints through Riemannian self-attention in hyperbolic space; 2) Relation-level alignment explicitly models pairwise spatial relationships between objects using the Information-Symplectic Relation Encoder (ISRE), which reformulates relation features through Fisher-Rao metric and Hamiltonian dynamics for uncertainty-aware geometrically consistent propagation; 3) Global-level alignment synthesizes discriminative global descriptors via the Spectral Manifold Transform (SMT) that extracts structural invariants through graph spectral analysis. This hierarchical alignment strategy progressively captures fine-grained to coarse-grained scene semantics, enabling robust cross-modal retrieval. Extensive experiments on the KITTI360Pose dataset demonstrate that SympLoc achieves a 19\% improvement in Top-1 recall@10m compared to existing state-of-the-art approaches.
\end{abstract}

\section{Introduction}
Understanding natural language instructions in city-scale 3D environments is critical for embodied AI and autonomous systems. Traditional visual place recognition (VPR) \cite{lowry2015visual} suffers from limited robustness against environmental shifts and strict viewpoint constraints. Text-to-point-cloud localization addresses this by directly correlating textual descriptions with 3D spatial signatures, eliminating physical proximity constraints. Existing methods (e.g., Text2Pos \cite{kolmet2022text2pos}, RET \cite{wang2023text}, Text2Loc \cite{xia2024text2loc}, MambaPlace \cite{shang2024mambaplace}) typically adopt a coarse-to-fine paradigm. However, their coarse retrieval stage predominantly relies on globally pooled descriptors in Euclidean space, leading to two fundamental limitations. First, {Severe Information Loss}: global pooling collapses discriminative instance-level and pairwise relational structures into a single vector, making it difficult to distinguish similar locations with partial semantic overlap. Second, {Hierarchical Structure Mismatch}: standard Euclidean attention treats all positions equally, failing to model the exponential, tree-like hierarchies inherent in real-world scene layouts (e.g., items $\rightarrow$ desk $\rightarrow$ room). While adopting the widely used coarse-to-fine paradigm, our core innovation lies in fundamentally redesigning the coarse retrieval stage. To overcome the aforementioned bottlenecks, we propose SympLoc, a multi-level cross-modal alignment strategy within the coarse stage. By progressively capturing scene semantics from fine-grained to coarse-grained, this strategy shifts the cross-modal representation learning from flat Euclidean spaces to structured geometric manifolds:\\
\indent {Instance-Level Alignment via Riemannian Instance Enhancer (RIE):} We embed object features into a hyperbolic Poincaré ball. This naturally captures the nested, tree-like hierarchies of urban scenes via geometry-respecting Riemannian self-attention, solving the structural mismatch problem.\\ 
\indent {Relation-Level Alignment via Information-Symplectic Relation Encoder (ISRE):} We explicitly model spatial relations using the Fisher-Rao metric to dynamically penalize linguistic ambiguity. Furthermore, symplectic Euler integration is applied to ensure volume-preserving, distortion-free relation propagation. \\
\indent {Global-Level Alignment via Spectral Manifold Transform (SMT):} To prevent information loss during pooling, we synthesize robust, permutation-invariant global descriptors utilizing Chebyshev graph spectral analysis without explicit eigen-decomposition.\\ 
\indent Extensive experiments on the KITTI360Pose dataset demonstrate that the proposed SympLoc significantly outperforms state-of-the-art methods. The main contributions can be summarized as follows: 1) We introduce the RIE for hyperbolic hierarchical modeling, which naturally captures the nested, tree-like structures of 3D scene layouts. 2) We propose the ISRE for uncertainty-aware relation alignment, ensuring geometrically consistent and distortion-free spatial propagation. 3) We develop the SMT for robust spectral global retrieval, extracting permutation-invariant structural descriptors without information loss.
\section{Related Work}
\subsection{Visual Place Recognition}
Traditional 2D localization techniques generally rely on feature aggregation mechanisms, such as the Vector of Locally Aggregated Descriptors (VLAD) \cite{arandjelovic2016netvlad} and Generalized Mean (GeM), to extract global descriptors from query frames, facilitating direct 2D feature matching. To bolster system reliability, recent studies have explored the inherent correlations between local descriptors and their corresponding cluster centers \cite{li20252}. Simultaneously, the field has seen a shift toward end-to-end Transformer-based architectures to produce more discriminative global embeddings \cite{wang2022transvpr,li2024feature}. Specialized frameworks have also been developed to tackle environmental extremes, including degraded lighting \cite{li2024toward} and drastic shifts in camera perspective \cite{berton2023eigenplaces}. Recent breakthroughs have further pushed the boundaries of VPR, scaling localization capabilities from urban environments to a continental magnitude \cite{lindenberger2025scaling}. Moreover, robust positioning is increasingly achieved through cross-modal retrieval, such as text-to-image \cite{shang2025bridging} or point-cloud-to-image \cite{xu2024c2l} paradigms, which provide critical fallbacks when direct visual input is compromised.
\subsection{LiDAR Place Recognition}
LiDAR-based Place Recognition (LPR) generates global descriptors either from unstructured point clouds or Bird’s-Eye View (BEV) representations \cite{lin2024rsbev}. Pioneering works such as PointNetVLAD \cite{uy2018pointnetvlad} integrated PointNet \cite{qi2017pointnet} with the NetVLAD layer to condense spatial points into compact vectors. Subsequent BEV-centric models, like BevPlace \cite{luo2023bevplace} and I2P-Rec \cite{zheng2023i2p}, have improved rotational robustness via geometric transforms and utilized multimodal projections to gather richer spatial cues.\\
\indent Alternative LPR strategies focus on the intrinsic topology of point clouds through 3D CNNs or attention-driven graphs. For instance, LPD-Net \cite{liu2019lpd} employs graph structures to represent local geometry, while DH3D \cite{du2020dh3d} and SOENet \cite{xia2021soe} emphasize high-fidelity pose refinement. To ensure embedding consistency during incremental learning, InCloud \cite{knights2022incloud} was introduced. Furthermore, Transformer-based LPR systems have gained traction for their ability to model long-range context; TransLoc3D \cite{xu2021transloc3d} utilizes self-attention for global awareness, whereas NDT-Transformer \cite{zhou2021ndt} and PPT-Net \cite{hui2021pyramid} process NDT-transformed data and multi-scale features through hierarchical structures. \\
\indent Efficiency in large-scale scenarios is often achieved via sparse 3D convolutions, as demonstrated by MinkLoc3D \cite{komorowski2021minkloc3d}. More recent advancements include LCDNet \cite{cattaneo2022lcdnet}, which adopts optimal transport theory for superior matching, and LoGG3D-Net \cite{vidanapathirana2022logg3d}, which utilizes a local consistency loss to ensure viewpoint-invariant descriptor discriminability.
\subsection{Cross Modal Localization}
The domain of text-to-point cloud localization was established by Text2Pos \cite{kolmet2022text2pos}, introducing a multi-stage coarse-to-fine pipeline. This was followed by RET \cite{wang2023text}, which leveraged attention mechanisms, and Text2Loc \cite{xia2024text2loc}, which utilized contrastive learning and a matching-free refinement phase. State Space Models (SSMs) were recently integrated into this task by MambaPlace \cite{shang2024mambaplace}. While GOTPR \cite{jung2025gotpr} improved efficiency through intermediate graph structures, its utility remains primarily in the coarse retrieval stage. \\
Current research also addresses practical bottlenecks like data scarcity and cross-modal ambiguity. IFRP-T2P \cite{wang2024instance} achieves instance-free positioning using query extractors combined with RowColRPA/RPCA for spatial awareness. Uncertainty modeling has emerged as a key theme: PMSH \cite{feng2025partially} manages partially matching submaps by propagating uncertainty into the coarse retrieval metric. Similarly, CMMLoc \cite{xu2025cmmloc} tackles incomplete textual descriptions through a Cauchy-Mixture-Model (CMM) framework, further refined by cardinal direction data and spatial consolidation.
\begin{figure*}
    \centering
    \includegraphics[width=0.92\linewidth]{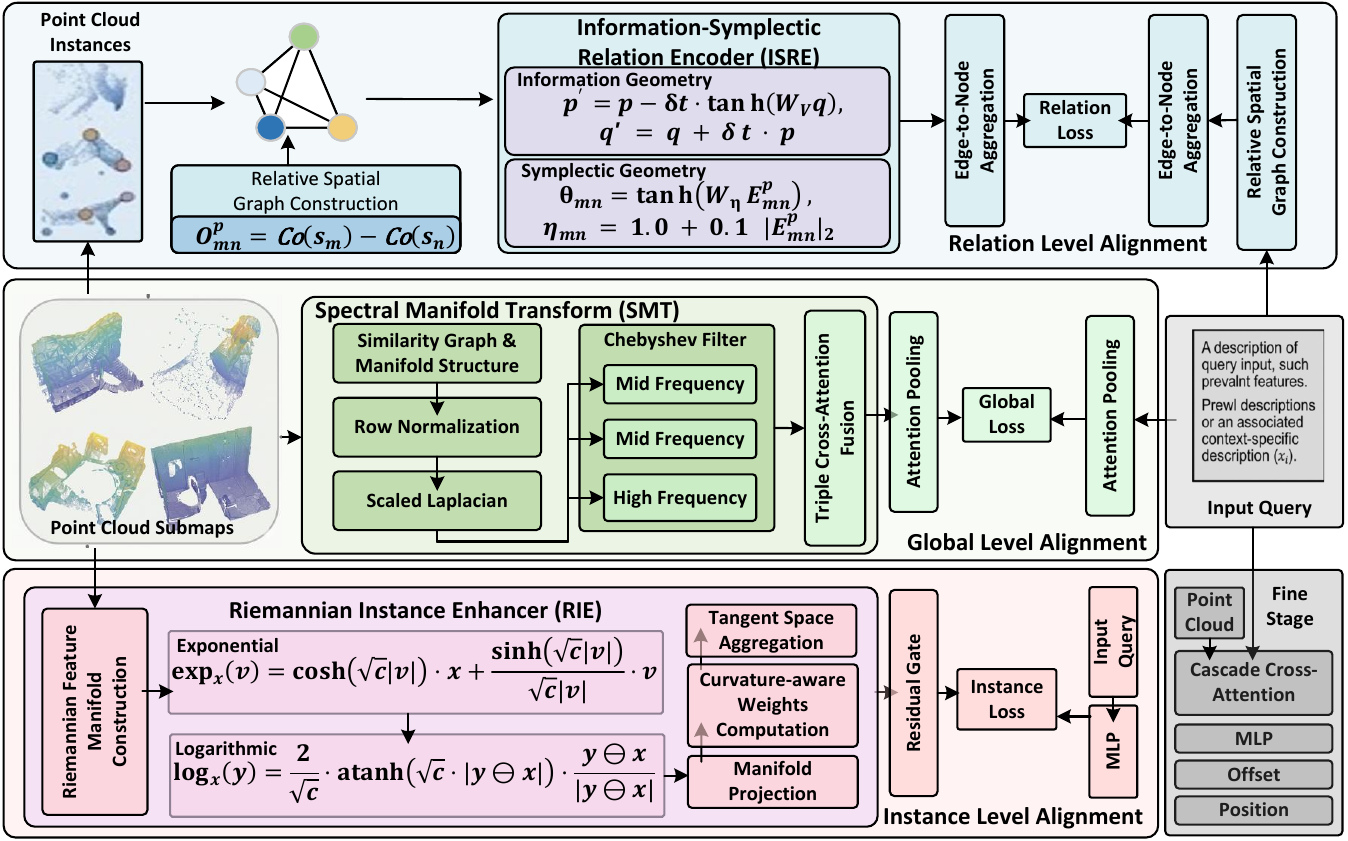}
    \vspace{-10pt}
    \caption{Overview of SympLoc, a coarse-to-fine framework for language-guided 3D point cloud localization. The framework comprises three complementary alignment modules in coarse stage: 1) Relation Level Alignment with the Information-Symplectic Relation Encoder (ISRE), which models pairwise spatial relationships through differential geometry; 2) Global Level Alignment with the Spectral Manifold Transform (SMT), which extracts structural invariants via Chebyshev spectral filtering; and 3) Instance Level Alignment with the Riemannian Instance Enhancer (RIE), which embeds features into hyperbolic space to capture hierarchical scene structures. The coarse stage retrieves the most relevant submap via multi-branch similarity aggregation, while the fine stage predicts the precise spatial location through cascaded cross-attention and MLP regression.}
    \label{Fig1}
\end{figure*}
\section{Methodology}
In this section, we present SympLoc, a novel framework for language-guided 3D point cloud localization. Given a query, SympLoc learns to predict the precise spatial location from a gallery of point cloud submaps.
\subsection{Task Formulation}
Let $\mathcal{G} = \{R_i\}_{i=1}^{N_c}$ represent the gallery of point cloud submaps, where each submap $R_i = \{s_k\}_{k=1}^{N_s}$ contains $N_s$ instances. The primary aim of this work is to determine the precise spatial location associated with a natural language query $Q_i$ (the $i$-th query in the dataset). Each query $Q_i$ comprises a sequence of $N_q$ relational descriptions, denoted by $\{d_j\}_{j=1}^{N_q}$. \\
\indent We tackle this task through a coarse-to-fine strategy. In the coarse stage, the most relevant point cloud submap is retrieved using the textual query. In the fine stage, we determine the exact position within the retrieved submap. \\
\indent As shown in Figure \ref{Fig1}, the main innovation of SympLoc lies in the coarse stage, where we decompose the cross-modal alignment into three complementary branches: 1) {Instance-level alignment} for modeling individual object semantics, 2) {Global-level alignment} for capturing holistic scene structure, and 3) {Relation-level alignment} for encoding pairwise spatial relationships. We detail the design of each branch in the following subsections. 

This entire process is optimized by minimizing the distance between the ground-truth position and the predicted position over the data distribution $\mathcal{D}_s$:
\begin{equation}
\min_{\Psi,\,\Phi} \ \mathbb{E}_{(L_{\text{gt}},\,Q_i) \sim \mathcal{D}_s}
\bigl\| L_{\text{gt}} - \Phi\!\left(Q_i, \tilde{R}_i\right) \bigr\|_2^2
\label{eq:task_obj_main}
\end{equation}
where $L_{\text{gt}} \in \mathbb{R}^2$ is the ground truth position. $\tilde{R}_i$ is the submap retrieved from $\mathcal{G}$ for the $i$-th query $Q_i$:
\begin{equation}
\tilde{R}_i = \arg\min_{R \in \mathcal{G}} \Delta \bigl( \Psi(Q_i), \Psi(R_i) \bigr)
\label{eq:task_obj_retrieval}
\end{equation}
Eq. \eqref{eq:task_obj_main} describes the fine stage, where the regressor $\Phi$ fuses the query with the retrieved submap $\tilde{R}_i$ to predict the final coordinates. Eq. \eqref{eq:task_obj_retrieval} describes the coarse stage: the retrieval function $\Psi$ embeds the query and each submap into a shared latent space, uses the Euclidean distance $\Delta(\cdot,\cdot)$ to compute similarity, and selects the nearest neighbor.
\subsection{Relation Level Alignment}
Given a query $Q_i$ with descriptions $\{d_j\}_{j=1}^{N_q}$ and a submap $R_i$ with instances $\{s_k\}_{k=1}^{N_s}$, we first extract their features. The descriptions are encoded via a frozen T5 encoder to obtain language features $\mathcal{T} = \{t_j\}_{j=1}^{N_q}$, and the point cloud instances are encoded via a frozen PointNet++ encoder to obtain point cloud features $\mathcal{X} = \{x_k\}_{k=1}^{N_s}$.
\subsubsection{Relative Spatial Graph Construction}
Natural language descriptions are inherently ambiguous, and object appearances may repeat within a scene. To address these challenges, we explicitly model the pairwise spatial relationships between instances, which serve as the most discriminative features for cross-modal grounding. For the point cloud modality, we construct a pairwise offset tensor $O^{p}_{mn} \in \mathbb{R}^{N_s \times N_s \times 3}$ to encode relative spatial information:
\begin{equation}
O^{p}_{mn} = \mathrm{Coord}(s_m) - \mathrm{Coord}(s_n)
\end{equation}
where $m, n \in \{1, \dots, N_s\}$ denote the instance indices, and $\mathrm{Coord}(\cdot)$ returns the instance centroid. $O^{p}_{mn}$ represents the geometric vector from instance $n$ to $m$.\\
\indent We fuse this geometric prior with semantic features to generate initial edge representations. For each instance pair $(m, n)$, the point cloud edge feature $E^{p}_{mn}$ is computed by concatenating the point cloud features with a projected geometric offset:
\begin{equation}
E^{p}_{mn} = \mathrm{MLP}_{\text{fuse}}\Bigl( \bigl[ x_m, x_n; \mathrm{MLP}_{\text{geo}}(O^{p}_{mn}) \bigr] \Bigr)
\end{equation}
For the textual modality, we generate a textual relation tensor $O^{t}_{mn}$ by concatenating the semantic embeddings of description pairs and projecting them through an MLP. Then we generate $E^{t}_{mn}$ following the same fusion procedure. This process maps implicit linguistic relations into a shared relational space, enabling alignment with point cloud geometric offset features.
\subsubsection{Information-Symplectic Relation Encoder (ISRE)}
Cross-modal spatial alignment requires discriminative features and geometrically consistent pairwise relations. A critical challenge arises from the inherent ambiguity of textual spatial descriptions compared to precise geometric cues: large absolute distances in point cloud relation features can dominate the embedding space, whereas textual relations encode only qualitative directional cues.\\
\indent To address this, we introduce the Information-Symplectic Relation Encoder (ISRE), which reformulates relation features through differential geometry. ISRE comprises two complementary components.\\
\indent \textbf{Information Geometry Layer:} \label{sec:info_geom}
We model each edge feature $E^{p}_{mn} \in \mathbb{R}^D$ as the natural parameter of an exponential family distribution on a statistical manifold $\mathcal{M}$ equipped with the Fisher-Rao metric. We project the relation embedding through a learnable transformation:
\begin{equation}
\boldsymbol{\theta}_{mn} = \tanh\!\bigl( W_\eta \, E^{p}_{mn} \bigr), \qquad
\eta_{mn} = 1.0 + 0.1 \, \|E^{p}_{mn}\|_2
\end{equation}
where $W_\eta \in \mathbb{R}^{D \times D}$ is a learnable projection and $\eta_{mn}$ is a precision parameter inversely related to the embedding norm. The Fisher-Rao distance between two relations $i$ and $j$ on the manifold is approximated as $d_{FR}(i,j) = \|\boldsymbol{\theta}_i - \boldsymbol{\theta}_j\|_2 / \sqrt{\eta}$. This uncertainty-aware metric naturally downweights noisy relations while preserving directional structure.\\
\indent \textbf{Symplectic Geometry Layer:} \label{sec:symplectic}
Hamiltonian mechanics provides a principled framework for structured information propagation. We decompose the natural parameter $\boldsymbol{\theta}_{mn}$ into position and momentum components $(\mathbf{q}, \mathbf{p})$ along the feature dimension: $\mathbf{q} = \boldsymbol{\theta}_{mn}[:, :D/2]$ and $\mathbf{p} = \boldsymbol{\theta}_{mn}[:, D/2:]$. The system evolves under a learnable potential $V(\mathbf{q}) = \frac{1}{2}\|\mathbf{W}_V \mathbf{q}\|_2^2$ with $\mathbf{W}_V \in \mathbb{R}^{D/2 \times D/2}$. The symplectic Euler update is:
\begin{equation}
\mathbf{p}' = \mathbf{p} - \delta t \cdot \tanh\!\bigl(\mathbf{W}_V \mathbf{q}\bigr), \qquad
\mathbf{q}' = \mathbf{q} + \delta t \cdot \mathbf{p}
\end{equation}
where $\delta t$ is a learnable time step. Crucially, the symplectic form $\omega(\mathbf{q}', \mathbf{p}') = \omega(\mathbf{q}, \mathbf{p})$ preserves phase-space volume, ensuring that information is neither created nor destroyed during propagation. This geometric prior encourages coherent relation updates across the spatial graph.\\
\indent The enhanced relation is obtained via a lightweight residual enhancement:
\begin{equation}
E^{p\star}_{mn} = \text{LN}\!\bigl( E^{p}_{mn} + \alpha_{\text{res}} \, \mathrm{concat}(\mathbf{q}', \mathbf{p}') \bigr)
\end{equation}
where $\alpha_{\text{res}}$ is a residual scaling factor and $\text{LN}(\cdot)$ denotes layer normalization. This bottleneck-then-reconstruct design keeps the module parameter-efficient ($\sim 33$K params for $D=256$), while the geometric priors (Fisher-Rao metric and symplectic dynamics) regularize the relation space without imposing strong inductive biases.\\
\indent \textbf{Edge-to-Node Aggregation:} \label{sec:edge_agg}
Pairwise relation features $E^{p\star} \in \mathbb{R}^{N_s \times N_s \times D}$ capture object interactions but cannot be directly aligned with textual descriptions. We compress them into compact node-level descriptors via attention-based aggregation:
\begin{equation}
\hat{x}_m = \sum_{n=1}^{N_s} \frac{\exp\!\bigl( E^{p\star}_{mn} \bigr)}{\sum_{k=1}^{N_s} \exp\!\bigl( E^{p\star}_{mk} \bigr)} \odot E^{p\star}_{mn}
\end{equation}
where the softmax is computed over the relational axis and $\odot$ denotes element-wise self-gating. This yields a spatial descriptor set $\hat{\mathcal{X}} = \{\hat{x}_m\}_{m=1}^{N_s}$ that aggregates each instance's contextual relations while remaining permutation-invariant. An identical aggregation is applied to $E^{t}_{mn}$ to obtain $\hat{\mathcal{T}} = \{\hat{t}_j\}_{j=1}^{N_q}$.
\medskip
\subsubsection{Relation Level Alignment Loss.}
Consider a batch of size $B$. Let $\hat{\mathcal{X}}_i$ and $\hat{\mathcal{T}}_j$ be the aggregated feature sets for point cloud submaps and descriptions. We define a bidirectional set-to-set similarity:
\begin{equation}
\mathcal{S}(\hat{\mathcal{X}}_i, \hat{\mathcal{T}}_j) = \frac{\exp(\lambda_{ij}^{\mathcal{X} \to \mathcal{T}}/\gamma)}{\sum_{b=1}^B \exp(\lambda_{ib}^{\mathcal{X} \to \mathcal{T}}/\gamma)} + \frac{\exp(\lambda_{ij}^{\mathcal{T} \to \mathcal{X}}/\gamma)}{\sum_{b=1}^B \exp(\lambda_{ib}^{\mathcal{T} \to \mathcal{X}}/\gamma)}
\label{eq:relation_similarity}
\end{equation}
where $\gamma$ is a temperature parameter and $\lambda_{ij}^{\mathcal{X} \to \mathcal{T}}$ represents the average of the maximum similarity scores between $\hat{\mathcal{X}}_i$ and $\hat{\mathcal{T}}_j$. The alignment loss is defined as:
\begin{equation}
\mathcal{L}_{\text{align}}(\hat{\mathcal{X}}_i, \hat{\mathcal{T}}_i) = -\frac{1}{B} \sum_{i,j=1}^B \mathbb{I}_{ij} (1 - \mathcal{J})^{\frac{1}{a}} \log \bigl( 1 - \mathcal{S}(\hat{\mathcal{X}}_i, \hat{\mathcal{T}}_j) \bigr)
\label{eq:relation_loss}
\end{equation}
where $\mathbb{I}_{ij}$ is an indicator function ($\mathbb{I}_{ij}=1$ for negative pairs, $0$ for positive pairs), $a$ is a learnable scale parameter, and $\mathcal{J}$ is a learnable weighted parameter.\\
\indent Textual descriptions are inherently ambiguous and asymmetric compared to precise geometric cues. Explicitly pulling positive pairs together may reinforce incomplete cross-modal alignments. Therefore, we only optimize the repulsion of negative pairs, using language primarily as exclusionary evidence. The total relation level loss is $\mathcal{L}_{RA} = \mathcal{L}_{\text{align}}(\hat{\mathcal{X}}, \hat{\mathcal{T}})$.
\subsection{Global Level Alignment}
In the coarse retrieval stage, we need to match the entire query with the submap, which requires a global descriptor that is discriminative and invariant to instance permutations. To this end, we propose the Spectral Manifold Transform (SMT), which extracts structural invariants from instance features through graph spectral analysis without explicit eigen-decomposition.\\
\indent Given the point cloud feature set $\mathcal{X} = \{x_k\}_{k=1}^{N_s}$ from the backbone, we construct an input feature matrix $\mathbf{X} \in \mathbb{R}^{N_s \times D}$ by stacking the $D$-dimensional features. We first characterize the manifold structure by computing a self-similarity adjacency matrix $\mathbf{A} \in \mathbb{R}^{N_s \times N_s}$:
\begin{equation}
A_{mn} = \exp\!\left(-\frac{\|\mathbf{x}_m - \mathbf{x}_n\|_2^2}{\tau}\right) + \delta_{mn}
\label{eq:similarity_graph}
\end{equation}
where $\tau > 0$ is a learnable temperature parameter controlling graph sparsity, and $\delta_{mn}$ is the Kronecker delta ensuring self-loops for centered aggregation. For spectral consistency, the matrix is row-normalized:
\begin{equation}
\widehat{A}_{mn} = \frac{A_{mn}}{\sum_{k=1}^{N_s} A_{mk} + \epsilon}
\end{equation}
where $\epsilon$ is a small constant for numerical stability. From $\widehat{\mathbf{A}}$, we derive the symmetric normalized Laplacian:
\begin{equation}
\mathbf{L} = \mathbf{I} - \mathbf{D}^{-\frac{1}{2}} \widehat{\mathbf{A}} \mathbf{D}^{-\frac{1}{2}}
\end{equation}
where $\mathbf{D}$ is the degree diagonal matrix with entries $D_{mm} = \sum_{n} \widehat{A}_{mn}$. To satisfy convergence criteria for Chebyshev approximation, we scale the Laplacian to the spectral radius interval $[-1, 1]$:
\begin{equation}
\widetilde{\mathbf{L}} = \frac{\mathbf{L}}{\max_{m,n}|L_{mn}| + \epsilon}
\label{eq:scaled_laplacian}
\end{equation}
where the denominator serves as an empirical estimate of the largest eigenvalue.\\
\indent We approximate the spectral filter using $K$-th order Chebyshev polynomials of the first kind, $T_k(\widetilde{\mathbf{L}})$, following the recursive definition:
\begin{equation}
\begin{aligned}
&T_0(\widetilde{\mathbf{L}}) = \mathbf{I}, \quad T_1(\widetilde{\mathbf{L}}) = \widetilde{\mathbf{L}}, \\
&T_{k+1}(\widetilde{\mathbf{L}}) = 2\widetilde{\mathbf{L}} \cdot T_k(\widetilde{\mathbf{L}}) - T_{k-1}(\widetilde{\mathbf{L}})
\end{aligned}
\end{equation}
\indent The SMT processes features through three parallel branches $b \in \{1, 2, 3\}$ to capture low, mid, and high-frequency structural patterns:
\begin{equation}
\mathbf{Y}^{(b)} = \sum_{k=0}^{K-1} \widehat{\beta}_k^{(b)} \, T_k(\widetilde{\mathbf{L}}) \cdot \mathbf{X}
\label{eq:chebyshev_filter}
\end{equation}
where learnable coefficients $\beta_k^{(b)}$ are normalized via softmax:
\begin{equation}
\widehat{\beta}_k^{(b)} = \frac{\exp(\beta_k^{(b)})}{\sum_{k'=0}^{K-1} \exp(\beta_{k'}^{(b)})}
\end{equation}
\indent These features are aggregated via a Triple Cross-Attention mechanism. We compute cross-attention weights $\Omega_1$ (between $\mathbf{Y}^{(3)}$ and $\mathbf{Y}^{(1)}$) and $\Omega_2$ (between $\mathbf{Y}^{(3)}$ and $\mathbf{Y}^{(2)}$). The final representation $\boldsymbol{\kappa}$ is obtained with the combined attention weight $\Omega_{\text{comb}}=\Omega_1\odot\Omega_2$:
\begin{equation}
\boldsymbol{\kappa} = \text{Attention}\bigl(\Omega_{\text{comb}}, \mathbf{Y}^{(3)}\bigr)
\end{equation}
The modulated feature $\boldsymbol{\kappa}$ is processed by a bidirectional GRU and a Transformer encoder to model long-range sequential dependencies. We extract the final global point descriptor $P^{g}_i \in \mathbb{R}^D$ using strided hidden-unit sampling for dimension reduction, while the language global descriptor $Q^{g}_i$ is generated via symmetric attention-pooling.\\
\indent We optimize global alignment using a contrastive similarity score $\mathcal{S}_{ij}$ between the $i$-th point cloud and $j$-th text query:
\begin{equation}
\mathcal{S}_{ij} = \frac{\exp\bigl((P^{g}_i)^\top Q^{g}_j/\gamma\bigr)}{\sum_{b=1}^B \exp\bigl((P^{g}_i)^\top Q^{g}_b/\gamma\bigr)} + \frac{\exp\bigl((Q^{g}_i)^\top P^{g}_j/\gamma\bigr)}{\sum_{b=1}^B \exp\bigl((Q^{g}_i)^\top P^{g}_b/\gamma\bigr)}
\label{eq:global_similarity}
\end{equation}
where $\gamma$ is the temperature hyperparameter and $B$ is the batch size. The global loss $\mathcal{L}_{\text{Global}}$ is defined as:
\begin{equation}
\mathcal{L}_{\text{Global}}= -\frac{1}{B} \sum_{i,j=1}^B \mathbb{I}_{ij} (1 - \mathcal{J})^{\frac{1}{a}} \log \bigl( 1 - \mathcal{S}_{ij} \bigr)
\label{eq:global_loss}
\end{equation}
where $\mathbb{I}_{ij}$ is an indicator function ($\mathbb{I}_{ij}=1$ for negative pairs, $0$ for positive pairs), $\mathcal{J}$ denotes the spatial overlap (IoU) between submaps to weigh the loss by proximity, and $a$ is a learnable scaling parameter controlling loss curvature.
\subsection{Instance Level Alignment}
While relation-level features capture pairwise interactions, individual instance descriptors still lack a structured representation of hierarchical relationships inherent in scene layouts. Specifically, instances form tree-like hierarchies (items on a desk to furniture in a room to the building itself), where standard self-attention assumes flat Euclidean geometry and fails to model the exponential sparsity of distant relationships.
\subsubsection{Riemannian Instance Enhancer} To address this, we propose the Riemannian Instance Enhancer (RIE), which embeds instance features into Hyperbolic space to naturally encode hierarchical structures. The hyperbolic metric expands space exponentially to accommodate sparse distant relationships while compressing dense near relationships, directly matching the geometric prior of scene layouts.\\
\begin{table*}
\centering
\caption{A comprehensive comparison between SympLoc and existing SOTA methods on the KITTI360Pose dataset.  (Best results are \textbf{bolded}, and second-best results are \underline{underlined}.)}
\label{table:compare_with_sota}
\resizebox{\linewidth}{!}{
\begin{tabular}{ccccccc}
\hline
\multirow{2}{*}{Methods} & \multicolumn{6}{c}{Localization Recall ($\epsilon < 5/10/15m$) $\uparrow$} \\ \cline{2-7} 
 & \multicolumn{3}{c}{Validation Set} & \multicolumn{3}{c}{Test Set} \\ \cline{2-7} 
 & k = 1 & k = 5 & k = 10 & k = 1 & k = 5 & k = 10 \\ \hline
Text2Pos (\textcolor{blue}{CVPR'22}) \cite{kolmet2022text2pos} & 0.14/0.25/0.31 & 0.36/0.55/0.61 & 0.48/0.68/0.74 & 0.13/0.20/0.30 & 0.33/0.42/0.49 & 0.43/0.61/0.65 \\
RET (\textcolor{blue}{AAAI'23}) \cite{wang2023text} & 0.19/0.30/0.37 & 0.44/0.62/0.67 & 0.52/0.72/0.78 & 0.16/0.25/0.29 & 0.35/0.51/0.56 & 0.46/0.65/0.71 \\
Text2Loc (\textcolor{blue}{CVPR'24}) \cite{xia2024text2loc} & 0.37/0.57/0.63 & 0.68/0.85/0.87 & 0.77/0.91/0.93 & 0.33/0.48/0.52 & 0.60/0.75/0.78 & 0.70/0.84/0.86 \\
IFRP-T2P (\textcolor{blue}{ACM MM’24}) \cite{wang2024instance}& 0.23/0.45/0.53 & 0.53/0.70/0.81 & 0.64/0.86/0.89 & 0.22/0.40/0.46 & 0.47/0.68/0.73 & 0.58/0.78/0.82 \\
MambaPlace (\textcolor{blue}{IROS’25})\cite{shang2024mambaplace} & \underline{0.45}/\underline{0.62}/\underline{0.68} & \underline{0.75}/\underline{0.89}/\underline{0.90} & \textbf{0.83}/\underline{0.94}/\underline{0.95} & 0.38/0.52/0.55 & 0.66/0.79/0.81 & 0.76/0.87/0.89 \\
CMMLoc (\textcolor{blue}{CVPR'25}) \cite{xu2025cmmloc}& 0.44/\underline{0.62}/\underline{0.68} & \underline{0.75}/0.88/\underline{0.90} & \textbf{0.83}/0.93/\underline{0.95} & \underline{0.39}/0.53/0.56 & 0.67/0.80/0.82 & 0.77/0.87/0.89 \\
PMSH (\textcolor{blue}{ICCV'25}) \cite{feng2025partially}& 0.42/\underline{0.62}/\underline{0.68} & \underline{0.75}/\underline{0.89}/\underline{0.90} & \textbf{0.83}/\underline{0.94}/\underline{0.95} & \underline{0.39}/\underline{0.55}/\underline{0.59} & \underline{0.68}/\underline{0.81}/\underline{0.83} & \underline{0.78}/\underline{0.89}/\underline{0.90} \\
SympLoc & \textbf{0.52/0.78/0.84} & \textbf{0.78/0.96/0.97} & \textbf{0.83/0.98/0.99} & \textbf{0.52/0.74/0.77} & \textbf{0.78/0.93/0.94} & \textbf{0.84/0.96/0.97} \\ \hline
\end{tabular}
}
\end{table*}
\indent \textbf{Riemannian Feature Manifold Construction:}
Instance relationships are non-linear: proximal instances share dense relationships while distal ones become exponentially sparse. To capture this structural prior, we embed the feature space into a Riemannian manifold $\mathcal{M}$ with a learnable constant curvature $c > 0$, equipped with the hyperbolic metric that provides superior capacity for encoding nested, hierarchical configurations.\\
\indent Given the instance feature matrix $\mathbf{V} \in \mathbb{R}^{N_s \times D}$, we first map Euclidean features to the manifold via a learnable scaling transformation:
\begin{equation}
\tilde{\mathbf{v}}_k = \frac{\mathbf{v}_k}{\|\mathbf{v}_k\|_2} \cdot \tanh(\zeta), \quad \tilde{\mathbf{v}}_k \in \mathcal{M}
\label{eq:manifold_projection}
\end{equation}
where $\zeta \in \mathbb{R}$ is a learnable scalar controlling geodesic scale, and $\tanh(\cdot)$ ensures bounded embeddings within the Poincar\'e ball of radius $\tanh(\zeta)$. To allow adaptive learning of the underlying geometric curvature, the curvature $c$ is parameterized as a softplus-activated learnable parameter:
\begin{equation}
c = \text{Softplus}(c_{\text{raw}}) + \epsilon, \quad c_{\text{raw}} \in \mathbb{R}
\label{eq:curvature_parameterization}
\end{equation}
where $\epsilon = 10^{-6}$ prevents numerical instability.\\
\indent \textbf{Riemannian Self-Attention Mechanism:} Standard attention mechanisms ignore space curvature, leading to structural distortion when processing hierarchical instance relationships. To model intrinsic relationships more faithfully, we propose a Riemannian self-attention (R-SA) mechanism. The core operations utilize the exponential map $\exp_x: T_x\mathcal{M} \rightarrow \mathcal{M}$ and the logarithmic map $\log_x: \mathcal{M} \rightarrow T_x\mathcal{M}$ to transition between the manifold and its tangent spaces, enabling geometry-respecting feature aggregation.
\medskip
\noindent\textbf{Exponential Map.} For a point $x \in \mathcal{M}$ and a tangent vector $v \in T_x\mathcal{M}$, the exponential map projects $v$ along the geodesic starting from $x$ back onto the manifold:
\begin{equation}
\exp_x(v) = \cosh(\sqrt{c} \|v\|) \cdot x + \frac{\sinh(\sqrt{c} \|v\|)}{\sqrt{c} \|v\|} \cdot v
\label{eq:exp_map}
\end{equation}
\medskip
\noindent\textbf{Logarithmic Map.} Conversely, the logarithmic map retrieves the tangent vector in $T_x\mathcal{M}$ connecting $x$ to $y$:
\begin{equation}
\log_x(y) = \frac{2}{\sqrt{c}} \cdot \operatorname{atanh}\!\left(\sqrt{c} \cdot \|y \ominus x\|\right) \cdot \frac{y \ominus x}{\|y \ominus x\|}
\label{eq:log_map}
\end{equation}
where $\ominus$ denotes the M\"obius subtraction, defining displacement in hyperbolic space:
\begin{equation}
x \ominus y = \frac{(1 + 2c\langle x, y \rangle + c\|y\|^2)x - (1 - c\|x\|^2)y}{1 + 2c\langle x, y \rangle + c^2\|x\|^2\|y\|^2}
\label{eq:mobius_subtraction}
\end{equation}
\indent \textbf{Riemannian Attention Computation.} Let $\mathbf{Q}, \mathbf{K}, \mathbf{V} \in \mathbb{R}^{N_s \times D}$ be projected from $\tilde{\mathbf{V}}$. The process is as follows:
\begin{itemize}
    \item \emph{Manifold Projection:} $\mathbf{Q}_{\mathcal{M}} = \phi_Q(\mathbf{Q})$, $\mathbf{K}_{\mathcal{M}} = \phi_K(\mathbf{K})$, $\mathbf{V}_{\mathcal{M}} = \phi_V(\mathbf{V})$.
    \item \emph{Curvature-aware Weights:} Compute $w_{mn}$ using manifold cosine similarity, which accurately measures proximity in non-Euclidean space:
    \begin{equation}
    w_{mn} = \frac{\exp\!\left(\frac{\langle \mathbf{q}_{\mathcal{M},m}, \mathbf{k}_{\mathcal{M},n} \rangle}{\sqrt{D}}\right)}{\sum_{n'=1}^{N_s} \exp\!\left(\frac{\langle \mathbf{q}_{\mathcal{M},m}, \mathbf{k}_{\mathcal{M},n'} \rangle}{\sqrt{D}}\right)}
    \label{eq:riemannian_attention_weights}
    \end{equation}
    \item \emph{Tangent Space Aggregation:} Aggregate values in the tangent space $T_{\mathbf{q}_{\mathcal{M},m}}\mathcal{M}$ before mapping back to the manifold:
    \begin{equation}
    \tilde{\mathbf{v}}_{\mathcal{M},m} = \exp_{\mathbf{q}_{\mathcal{M},m}}\!\left(\sum_{n=1}^{N_s} w_{mn} \cdot \log_{\mathbf{q}_{\mathcal{M},m}}(\mathbf{v}_{\mathcal{M},n})\right)
    \label{eq:riemannian_attention_aggregate}
    \end{equation}
\end{itemize}
\indent The enhanced feature $\mathbf{V}_{\text{RSA}} = \psi(\tilde{\mathbf{V}}_{\mathcal{M}})$ is returned to Euclidean space via normalization $\psi(\cdot)$. A residual connection with a learnable gate $\beta \in [0, 1]$ preserves original features:
\begin{equation}
\mathbf{V}^* = \beta \cdot \mathbf{V}_{\text{RSA}} + (1 - \beta) \cdot \mathbf{V}
\label{eq:residual_connection}
\end{equation}
\subsubsection{Instance Level Alignment Loss. }
After Riemannian enhancement, we obtain the refined point cloud instance descriptors $\mathcal{V}^* = \{\mathbf{v}^*_m\}_{m=1}^{N_s}$. These are projected alongside text instance features $\mathcal{T} = \{t_j\}_{j=1}^{N_q}$ via $\mathrm{MLP}_{\text{inst}}$ to generate aligned descriptors $\mathcal{V}'$ and $\mathcal{T}'$. The resulting alignment is optimized using a contrastive loss $\mathcal{L}_{\text{IA}} = \mathcal{L}_{\text{align}}(\mathcal{V}', \mathcal{T}')$, ensuring point cloud instances are deeply aligned with their corresponding linguistic semantics within a shared geometric framework.
\subsection{Fine Stage} 
In the coarse retrieval phase, the proposed framework employs a tripartite architecture where each branch is optimized independently. We aggregate the individual similarity scores from these branches through summation to generate a unified composite score. This integrated metric is subsequently used to rank and select the most relevant candidate submaps for the ensuing fine-grained localization stage. \\
\indent For the final fine-grained localization within the retrieved submap, SympLoc directly adopts the well-established fine-stage pipeline from Text2Loc. Specifically, we employ a cascaded cross-attention mechanism to deeply fuse the geometry-enhanced point cloud features with the aligned text features. A standard multi-layer perceptron (MLP) regressor is then used to predict the relative spatial offset with respect to the submap anchor point, from which the precise spatial coordinates are directly obtained.

\begin{table}
\centering
\caption{Text to point cloud submap retrieval accuracy of SympLoc and SOTA methods in the coarse stage on the KITTI360Pose dataset. SympLoc (global) denotes the variant of our method that only utilizes the global branch in the coarse stage.}
\label{table:compare_coarse_stage}
\resizebox{\linewidth}{!}{
\begin{tabular}{ccccccc}
\hline
{Methods} & \multicolumn{6}{c}{Submap Retrieval Recall $\uparrow$} \\ \cline{2-7}
& \multicolumn{3}{c}{Validation Set} & \multicolumn{3}{c}{Test Set} \\ \cline{2-7}
& \multicolumn{1}{c}{k = 1} & \multicolumn{1}{c}{k = 3} & \multicolumn{1}{c}{k = 5} & \multicolumn{1}{c}{k = 1} & \multicolumn{1}{c}{k = 3} & k = 5 \\ \hline
Text2Pos & \multicolumn{1}{c}{0.14} & \multicolumn{1}{c}{0.28} & \multicolumn{1}{c}{0.37} & \multicolumn{1}{c}{0.12} & \multicolumn{1}{c}{0.25} & 0.33 \\
RET & \multicolumn{1}{c}{0.18} & \multicolumn{1}{c}{0.34} & \multicolumn{1}{c}{0.44} & \multicolumn{1}{c}{0.15} & \multicolumn{1}{c}{0.29} & 0.37 \\
Text2Loc & \multicolumn{1}{c}{0.31} & \multicolumn{1}{c}{0.54} & \multicolumn{1}{c}{0.64} & \multicolumn{1}{c}{0.28} & \multicolumn{1}{c}{0.49} & 0.58 \\
IFRP-T2P & \multicolumn{1}{c}{0.24} & \multicolumn{1}{c}{0.46} & \multicolumn{1}{c}{0.57} & \multicolumn{1}{c}{0.23} & \multicolumn{1}{c}{0.39} & 0.48 \\
MambaPlace & \multicolumn{1}{c}{0.35} & \multicolumn{1}{c}{0.61} & \multicolumn{1}{c}{0.72} & \multicolumn{1}{c}{0.31} & \multicolumn{1}{c}{0.53} & 0.62 \\
CMMLoc & \multicolumn{1}{c}{0.35} & \multicolumn{1}{c}{0.61} & \multicolumn{1}{c}{0.73} & \multicolumn{1}{c}{0.32} & \multicolumn{1}{c}{0.53} & 0.63 \\
PMSH & \multicolumn{1}{c}{0.37} & \multicolumn{1}{c}{0.63} & \multicolumn{1}{c}{0.73} & \multicolumn{1}{c}{0.34} & \multicolumn{1}{c}{0.56} & 0.65 \\ \hline
SympLoc (Global)
& \multicolumn{1}{c}{\underline{0.40}} & \multicolumn{1}{c}{\underline{0.65}} & \multicolumn{1}{c}{\underline{0.77}} & \multicolumn{1}{c}{\underline{0.39}} & \multicolumn{1}{c}{\underline{0.62}} & \underline{0.71} \\
SympLoc & \multicolumn{1}{c}{\textbf{0.54}} & \multicolumn{1}{c}{\textbf{0.81}} & \multicolumn{1}{c}{\textbf{0.89}} & \multicolumn{1}{c}{\textbf{0.51}} & \multicolumn{1}{c}{\textbf{0.75}} & \textbf{0.82} \\ \hline
\end{tabular}}
\end{table}
\begin{table}
\centering
\caption{Ablation study investigating the contributions of the three independent branches, Relation, Instance, and Global—within the coarse retrieval framework.}
\label{table:ablation_coarse_loss}
\resizebox{1\linewidth}{!}{ 
\begin{tabular}{ccc|ccc} % 
\hline
\multicolumn{3}{c|}{Methods} & \multicolumn{3}{c}{Retrieval Recall $\uparrow$} \\ \cline{1-6} 
Relation & Instance & Global & k=1 & k=3 & k=5 \\ \hline
\textbf{\checkmark} & & & 0.46 & 0.70 & 0.78 \\ 
& \textbf{\checkmark} & & 0.48 & 0.72 & 0.79 \\
& & \textbf{\checkmark} & 0.39 & 0.62 & 0.71 \\ 
\textbf{\checkmark} & \textbf{\checkmark} & & 0.50 & 0.74 & 0.81 \\ 
\textbf{\checkmark} & & \textbf{\checkmark} & 0.48 & 0.72 & 0.80 \\ 
& \textbf{\checkmark} & \textbf{\checkmark} & 0.49 & 0.73 & 0.80 \\ 
\textbf{\checkmark} & \textbf{\checkmark} & \textbf{\checkmark} & \textbf{0.51} & \textbf{0.75} & \textbf{0.82} \\ \hline
\end{tabular}}
\end{table}
\begin{table}
\centering
\caption{Ablation study on core innovations of SympLoc. We independently remove RIE, ISRE, and SMT to verify their contributions to the coarse-stage text-to-point cloud submap retrieval on the test set.}
\label{table:ablation}
\resizebox{\linewidth}{!}{
\begin{tabular}{ccccccc}
\hline
{}{}{Methods} & \multicolumn{6}{c}{Submap Retrieval Recall $\uparrow$} \\ \cline{2-7}
& \multicolumn{3}{c}{Validation Set} & \multicolumn{3}{c}{Test Set} \\ \cline{2-7}
& \multicolumn{1}{c}{k = 1} & \multicolumn{1}{c}{k = 3} & \multicolumn{1}{c}{k = 5} & \multicolumn{1}{c}{k = 1} & \multicolumn{1}{c}{k = 3} & k = 5 \\ \hline
w/o RIE & \multicolumn{1}{c}{0.52} & \multicolumn{1}{c}{0.78} & \multicolumn{1}{c}{0.86} & \multicolumn{1}{c}{0.49} & \multicolumn{1}{c}{0.73} & 0.80 \\
w/o ISRE & \multicolumn{1}{c}{0.53} & \multicolumn{1}{c}{0.79} & \multicolumn{1}{c}{0.87} & \multicolumn{1}{c}{0.49} & \multicolumn{1}{c}{0.71} & 0.81 \\
w/o SMT & \multicolumn{1}{c}{0.51} & \multicolumn{1}{c}{0.77} & \multicolumn{1}{c}{0.85} & \multicolumn{1}{c}{0.48} & \multicolumn{1}{c}{0.72} & 0.79 \\
SympLoc & \multicolumn{1}{c}{\textbf{0.54}} & \multicolumn{1}{c}{\textbf{0.81}} & \multicolumn{1}{c}{\textbf{0.89}} & \multicolumn{1}{c}{\textbf{0.51}} & \multicolumn{1}{c}{\textbf{0.75}} & \textbf{0.82} \\ \hline
\end{tabular}}
\end{table}
\section{Experiment}
\subsection{Experiment Setup}
\indent We conduct our evaluations on the KITTI360Pose benchmark, which spans a 15.51 km$^2$ urban area across nine distinct sequences. Following the standard protocol, the dataset is divided into training, validation, and testing sets, containing 28,689, 3,187, and 11,505 query-submap pairs, respectively. All models are implemented using PyTorch and executed on a workstation equipped with a 128-core AMD EPYC CPU and an NVIDIA RTX V100 (32 GB) GPU.\\
\indent For the coarse retrieval phase, SympLoc is trained for 20 epochs with a batch size of 64 and a learning rate of 0.0005, yielding a 256-dimensional global descriptor. In the subsequent fine-stage localization, we train the network for 100 epochs using a batch size of 32 and a reduced learning rate of 0.0003. These hyperparameter configurations are consistent with established baselines on this dataset to ensure a fair comparison.
\subsection{Overall Performance}
We first evaluate SympLoc against six state-of-the-art methods on the KITTI360Pose benchmark. As shown in Table~\ref{table:compare_with_sota}, SympLoc substantially outperforms all baselines across all metrics on both validation and test sets. Notably, on the test set Top-1 localization @10m, SympLoc achieves {0.74}, yielding an absolute improvement of 0.19 over MambaPlace (0.55) and CMMLoc (0.53), the two strongest prior methods. Even at the stricter @5m threshold—where accurate localization is more challenging—SympLoc still delivers a notable absolute gain of 0.13, reaching 0.52 compared to 0.39 for the best baseline. These results demonstrate that SympLoc's multi-level alignment strategy enables more precise localization than methods relying on flat Euclidean representations.\\
\indent The coarse-stage submap retrieval results in Table~\ref{table:compare_coarse_stage} reveal the source of SympLoc's success. By decomposing the retrieval problem into instance, relation, and global alignment, SympLoc captures scene structure at multiple granularities, leading to substantially better candidate selection. On the test set, SympLoc achieves Top-1 retrieval of {0.51} and Top-5 of {0.82}, achieving absolute gains of 0.17 and 0.17, respectively, over the best prior method PMSH (0.34/0.65). This confirms that the fundamental bottleneck in text-to-point-cloud localization lies in the coarse retrieval stage, and that multi-level structural alignment effectively addresses the information loss inherent in global pooling. \\
\indent To demonstrate that our performance gains are not solely derived from the multi-level architecture, we directly compare our global branch against existing baselines, which are entirely based on global descriptors. Even when operating with the Global branch alone, SympLoc achieves a Top-1 recall of 0.39 on the test set, yielding an absolute improvement of 0.05 over the best prior method PMSH (0.34). At Top-5, this global-only variant reaches 0.71, exceeding PMSH (0.65) by 6 percentage points. This is particularly significant because it isolates the contribution of the spectral global descriptor from the other branches: even without the complementary hierarchical cues from RIE or relational signals from ISRE, SMT's graph spectral analysis and Triple Cross-Attention mechanism already produce global descriptors with powerful representational capabilities. This confirms that extracting permutation-invariant structural invariants through Chebyshev spectral filtering is highly effective in capturing global scene features.
\subsection{Ablation for Three branches}
To verify that each branch contributes independently to retrieval quality, we progressively add branches and measure retrieval recall. As shown in Table~\ref{table:ablation_coarse_loss}, all three branches provide meaningful gains when used alone (Top-1: 0.39--0.48), and combining any two branches yields further improvement. The full model with all three branches achieves Top-1 recall of {0.51}, outperforming any single branch by 3--12 percentage points and any two-branch combination by 1 to 2 percentage points.\\
\begin{figure*}
    \centering
    \includegraphics[width=0.92\linewidth]{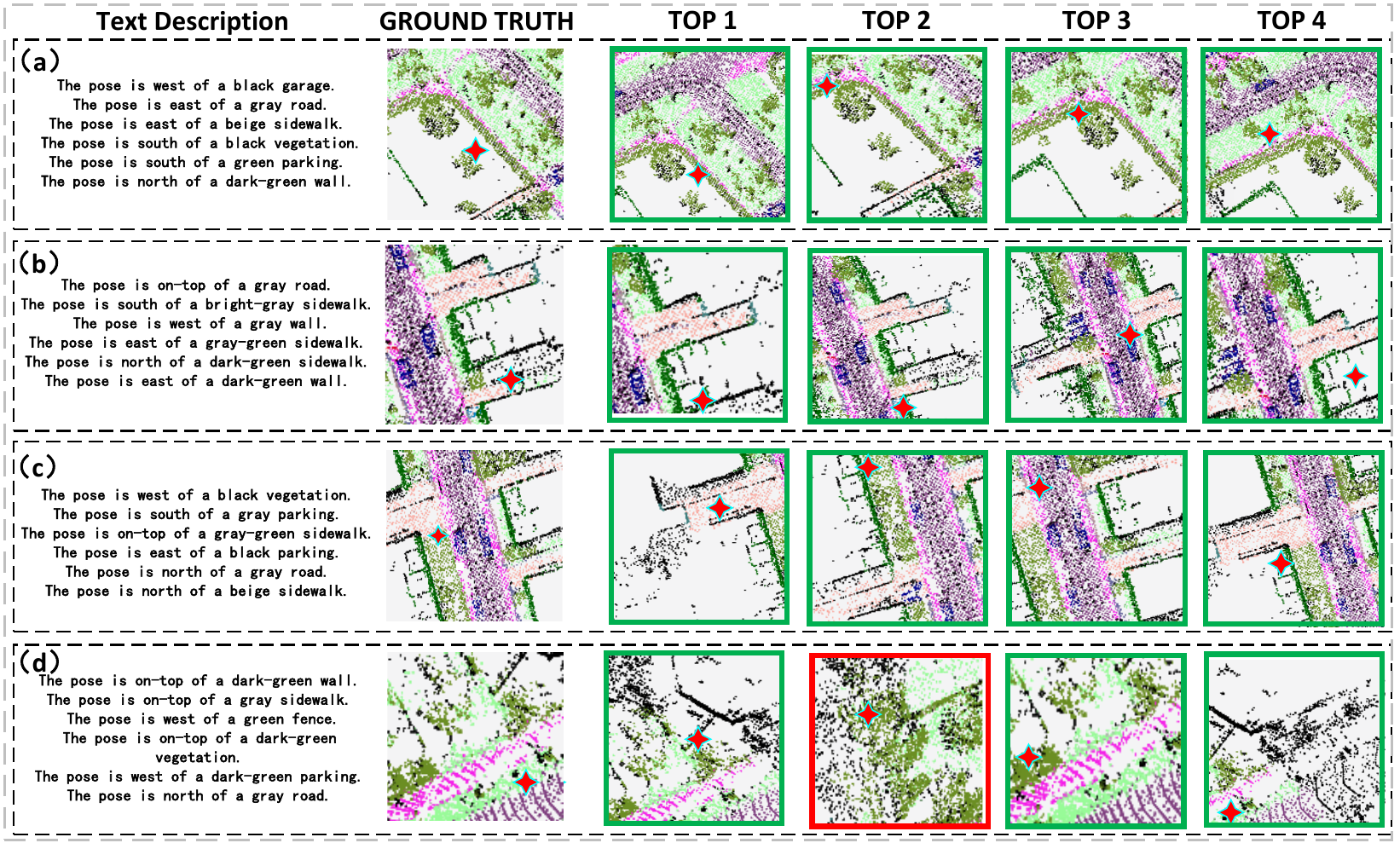}
    \vspace{-10pt}
    \caption{In this visualization analysis experiment, we present the textual descriptions, the point clouds of the actual location (ground truth), and the top K 
retrieved point-cloud submaps. We mark the retrieved position in each submap with a star symbol. If the Euclidean distance between the retrieved coordinates 
and the actual coordinates is within 15 meters, we consider this retrieval correct and highlight the submap with a green box. Incorrect submaps are marked 
with red boxes.}
    \label{Figure2}
\end{figure*}
\indent Several patterns emerge from this analysis. First, the {Instance} branch alone achieves the highest single-branch Top-1 recall (0.48), suggesting that hierarchical scene structure captured by RIE's hyperbolic embedding is the most discriminative signal for cross-modal matching. Second, the {Relation} branch (0.46) outperforms the {Global} branch (0.39) by a substantial margin, confirming that pairwise spatial relationships are more informative than pooled global descriptors for distinguishing between semantically similar candidates. Third, the {Global} branch, while weakest in isolation, contributes essential complementary information when combined with other branches---the full model outperforms the Relation+Instance combination by 1 percentage point at Top-1. This indicates that spectral structural invariants captured by SMT encode information orthogonal to both hierarchical and relational cues.\\ 
\indent The additivity of branch contributions validates SympLoc's architectural decision to optimize three independent alignment objectives and aggregate their similarity scores. The diminishing returns from adding the third branch (from Relation+Instance to full) reflect some redundancy across levels, yet each branch captures a distinct geometric or semantic aspect that no other branch can fully substitute.
\subsection{Ablation for Key Components}
We further ablate the three core modules, RIE, ISRE, and SMT, to isolate their individual contributions within their respective branches. As shown in Table~\ref{table:ablation}, removing any single module degrades performance, confirming that all three innovations are essential to SympLoc's superiority.\\
\indent Removing {RIE} (w/o RIE) reduces Top-1 retrieval from 0.51 to 0.49 on the test set. This drop is most pronounced at Top-3 (0.75 to 0.73) and Top-5 (0.82 to 0.80), indicating that hyperbolic hierarchical modeling primarily improves recall by helping distinguish submaps that share similar instance compositions but differ in structural organization. Without RIE, the model cannot represent the exponential sparsity of distant instance relationships, leading to conflation of hierarchically distinct scenes.\\
\indent Removing {ISRE} (w/o ISRE) causes the largest Top-3 drop (0.75 to 0.71) on the test set. This confirms that Fisher-Rao uncertainty-aware metric and symplectic dynamics are critical for modeling pairwise spatial relations. The ISRE's ability to dynamically downweight noisy relations while preserving directional structure enables more reliable cross-modal alignment of spatial descriptions. Without this module, textual ambiguity and geometric noise corrupt relation features, degrading retrieval, especially in the mid-rank range where candidates are plausible but not correct.\\
\indent Removing {SMT} (w/o SMT) reduces Top-1 from 0.51 to 0.48 and Top-5 from 0.82 to 0.79. The Chebyshev spectral filter's multi-frequency decomposition captures structural patterns at different resolutions, while the triple cross-attention mechanism synthesizes permutation-invariant global descriptors. Without SMT, the global descriptor loses sensitivity to manifold structure, making it harder to differentiate submaps that have similar instance distributions but distinct topological arrangements.\\
\indent It is worth noting that even after removing any individual module, the model still substantially outperforms the baseline method PMSH, demonstrating that SympLoc’s multi-level design achieves robust performance via complementary feature representations, rather than relying on a single dominant component.
\subsection{Visualization Results}
Through additional visualization analysis, we identify representative retrieval results from text queries, as shown in Figure \ref{Figure2}. For the majority of queries, SympLoc effectively retrieves scene candidates that align with the textual spatial descriptions, demonstrating its robust multi-level alignment capability. Notably, in cases such as (a) and (b), the Top-1 retrieved submaps exhibit high structural similarity with the ground truth, validating the effectiveness of our hierarchical scene modeling. However, in challenging cases such as (d), when textual descriptions contain ambiguous spatial relations, the Top-2 candidate may correspond to geographically distant locations with similar visual appearances.  Despite this inherent ambiguity, SympLoc still maintains high retrieval accuracy—the remaining three candidates (Top-1, Top-3, and Top-4) are all correctly localized—by leveraging the ISRE's uncertainty-aware Fisher-Rao metric and SMT's topology-invariant embeddings.
\section{Conclusion}
In this paper, we propose SympLoc, a novel coarse-to-fine localization framework with multi-level cross-modal alignment for language-guided 3D point cloud understanding. The core innovation lies in redesigning the coarse retrieval stage through three complementary alignment levels: Riemannian Instance Enhancer (RIE) for hyperbolic hierarchical modeling, Information-Symplectic Relation Encoder (ISRE) for uncertainty-aware geometrically consistent propagation, and Spectral Manifold Transform (SMT) for permutation-invariant global retrieval. Extensive experiments on KITTI360Pose demonstrate that SympLoc significantly outperforms state-of-the-art methods, achieving a 19\% improvement in Top-1 recall@10m. Future work includes extending SympLoc to larger-scale environments and exploring finer-grained temporal consistency for dynamic scene understanding.

\bibliographystyle{IEEEtran}
\bibliography{IEEEabrv,Bibliography}
\end{document}